\documentclass[preprint,review, 11pt, nonatbib]{elsarticle}
\usepackage[margin=1in]{geometry}
\usepackage{url}
\usepackage{amsmath,amssymb,amsfonts,latexsym, amsthm}
\usepackage{mathtools}
\usepackage{array}
\usepackage{xcolor}
\usepackage{graphicx}
\usepackage{caption}
\usepackage{subcaption}
\usepackage{threeparttable} 
\usepackage{booktabs}
\usepackage{multirow}
\usepackage{enumitem}
\usepackage{todonotes}
\usepackage{soul}
\usepackage[ruled,vlined,linesnumbered]{algorithm2e}
\usepackage{multicol}
\usepackage[hidelinks]{hyperref}
\usepackage{apacite}  
\usepackage{tabularx}  
\usepackage{float}
\usepackage{microtype}
\newcommand{\TS}{\text{TS}}
\newcommand{\ABS}{\text{ABS}}
\newcommand{\PATH}{\text{PATH}}
\newcommand{\ms}[2]{#1\,$\pm$\,#2} 

\journal{IISE Transactions on Healthcare Systems Engineering}
\graphicspath{{Figs/}}

\begin{document}

\begin{frontmatter}

\title{Multi-Modal Machine Learning for Breast Cancer Recurrence Prediction}

\author[1]{Jiahao Shao}
\author[1]{Xudong Wang}
\author[1]{Anam Nawaz Khan}
\author[2]{Christopher Brett}
\author[1]{Xueping Li}
\author[1]{Bing Yao\footnote{Corresponding author: byao3@utk.edu}}

\address[1]{Department of Industrial \& Systems Engineering, The University of Tennessee, Knoxville, TN 37996, USA}
\address[2]{The University of Tennessee Medical Center, Knoxville, TN 37920, USA}

\begin{abstract}
    Breast cancer recurrence, a leading cause of long-term mortality among survivors, requires timely and accurate risk assessment to guide follow-up care and treatment planning. Traditional predictive models, often limited to either structured or unstructured data alone, struggle to capture the full clinical context. This study examines the impact of integrating multi-modal clinical data, including treatment records, pathology reports, and clinician notes, on recurrence prediction. By integrating a rule-based regular expression extraction mechanism with a rigorous precedence-based conflict reconciliation strategy, our approach effectively recovers definitive tumor characteristics from free-text pathology narratives to augment structured records. We also benchmark performance against commonly used feature sets from prior breast cancer studies to assess the added value of multi-modal integration. Single-source and multi-modal inputs are evaluated across a range of machine learning models. Results show that multi-modal integration consistently improves predictive accuracy compared to single-modal methods.
\end{abstract}

\begin{keyword}
   Electronic health records, Multi-modal learning, Breast cancer, Regular expression extraction, Data harmonization
\end{keyword}

\end{frontmatter}

\section{Introduction}
    Breast cancer is the most commonly diagnosed malignancy in women worldwide and remains a leading cause of cancer mortality~\cite{kim2025global,qian2025monte}. In 2020, an estimated 2.3 million new cases and 685,000 deaths were recorded globally, representing 24.5\% of all new cancers and 15.5\% of all cancer deaths among women~\cite{sung2021global}. Despite significant advances in screening and systemic therapy, recurrence continues to drive long-term morbidity~\cite{schwarz2025prediction}. Particularly in hormone receptor–positive disease, distant recurrence risks persist well beyond the initial treatment window, with cumulative risks exceeding 20\% by 20 years after diagnosis~\cite{pan201720}. Accurate recurrence risk assessment is therefore critical for guiding surveillance intensity, informing adjuvant treatment decisions, and supporting individualized survivorship care.
    
    Current prognostic modeling largely relies on clinicopathologic variables defined by the American Joint Committee on Cancer (AJCC), including tumor-node-metastasis (TNM) staging, histological grade, and biomarker status, specifically estrogen receptor (ER), progesterone receptor (PR), and human epidermal growth factor receptor 2 (HER2)~\cite{giuliano2017breast, amin2017ajcc}. Professional guidelines from the American Society of Clinical Oncology/College of American Pathologists (ASCO/CAP) standardize the assessment of these markers, which directly influence staging and therapy selection~\cite{allison2020estrogen, wolff2023human}. While Electronic Health Records (EHRs) provide a rich data source for such modeling, real-world clinical data are characterized by heterogeneity and fragmentation~\cite{kaur2025enhancing}. Previous studies relying solely on structured data often lack the nuance contained in free-text narratives. For example, \citeA{alzu2021predicting} utilized structured EHR variables to predict recurrence but noted limitations in capturing granular tumor characteristics often buried in free-text notes. Conversely, approaches leveraging unstructured text have demonstrated the value of narrative information. \citeA{gonzalez2023machine}, for instance, showed that incorporating features extracted from clinical notes significantly improved predictive discrimination relative to baselines that relied exclusively on structured codes. 
    
    Recent multi-modal breast cancer recurrence models have achieved strong predictive performance by integrating imaging, clinical, and molecular signals through end-to-end fusion architectures. For example, Yao et al.~\cite{yao2022icsda} proposed an ICSDA framework integrating pathological, clinical, and gene-expression data, and Zhang et al.~\cite{zhang2025multimodal} developed a multimodal deep learning model linking recurrence prediction with Oncotype DX risk. These studies demonstrate the value of combining heterogeneous evidence sources through latent feature fusion and representation learning. However, most existing frameworks assume that the underlying clinical variables are already reliable and consistently curated across sources, an assumption that is often violated in routine EHR environments.    
    
    A significant challenge in leveraging EHRs for recurrence prediction is the discordance between data sources. Routine clinical data are frequently fragmented across registries, treatment summaries, and narrative documents, leading to incompleteness and inconsistency~\cite{weiskopf2013methods, ma2025large}. While synoptic pathology reporting improves consistency relative to free-text~\cite{sluijter2016effects, renshaw2018synoptic}, and registry linkage enhances variable capture~\cite{charlton2022cancer}, critical attributes such as histologic grade, Ki-67, and lymphovascular invasion (LVI) often suffer from block-wise missingness in structured fields. This reduces the effective sample size and degrades clinician trust. Furthermore, manual data entry introduces transcription errors; definitive information often originates in pathology reports but is prone to errors when transcribed into downstream systems~\cite{weiskopf2013methods, renshaw2018synoptic,wang2024rule}.
    
    To address these challenges, this paper proposes a multi-modal data harmonization framework for breast cancer recurrence prediction. Using data from the University of Tennessee Medical Center (UTMC), we harmonize three routinely available sources, i.e., treatment summaries ($\TS$), registry abstracts ($\ABS$), and pathology reports ($\PATH$), into a unified and learning-ready format. Our approach utilizes rule-based regular expression extractors to recover high-fidelity prognostic variables from pathology narratives and applies precedence-based logic to reconcile discrepancies across sources. By quantifying the impact of this fusion on data completeness and benchmarking performance across a range of machine learning models, 
    we demonstrate that clinically faithful multi-source integration yields superior predictive discrimination compared to single-source baselines.

\section{Research Background}

    \subsection{Structured EHR-based Prediction Models}
    The widespread adoption of EHRs \cite{wang2026muse,wang2024multi} has facilitated the development of automated decision support tools for oncology. Traditional machine learning methods, including logistic regression (LR), support vector machines (SVMs), and random forests (RF), have been extensively applied to predict breast cancer recurrence using structured clinical variables~\cite{jiang2025deep, ahmad2013using, al2018breast, lu2023predictive, hosmer2013applied, hearst1998support, breiman2001random}. 
    Early studies established the feasibility of applying machine learning to postoperative recurrence risk prediction but were constrained by limited sample sizes and methodological scope. For instance, \citeA{kim2016nomogram} developed a na\"{\i}ve Bayesian classifier using data from 679 post-surgical patients, integrating clinicopathologic variables to achieve approximately 80\% predictive accuracy, surpassing guideline-based risk stratification. This work extended earlier investigations by \citeA{kim2012development}, which employed SVMs to demonstrate that data-driven models could improve upon expert-derived scoring systems. 

   More recent studies have addressed these limitations by leveraging larger, prospectively collected cohorts and more expressive modeling techniques. Notably, \citeA{lou2020machine} conducted a comparative evaluation of multiple machine learning algorithms for 10-year recurrence prediction within a prospective cohort that incorporated treatment quality indicators, demonstrating consistently superior risk stratification relative to traditional approaches. Similarly, \citeA{zuo2023machine} applied gradient boosting and deep neural network models to structured EHR data, achieving improved discrimination and calibration in recurrence risk estimation.  Systematic reviews indicate that ensemble methods and deep learning architectures generally deliver the strongest predictive performance, with reported area under the receiver operating characteristic curve (AUROC) values approaching 0.9 in selected patient subgroups~\cite{el2023evolution}. However, most existing models implicitly treat structured EHR variables as ground truth, assuming both completeness and correctness of the recorded data. In real-world oncology practice, this assumption is frequently violated: EHR data are often fragmented across encounters, affected by systematic missingness, and susceptible to transcription and coding errors, all of which can introduce bias and degrade model reliability~\cite{carrell2014using}.

    \subsection{Clinical Text and Registry-Based Models}
    To overcome the limitations of structured data, recent research has increasingly leveraged unstructured clinical text and cancer registry data. Natural Language Processing (NLP) techniques have been employed to extract rich phenotypic information from oncology notes and pathology reports, effectively recovering recurrence outcomes that are incompletely coded in structured fields. For instance, \citeA{sanyal2021weakly} developed a weakly supervised deep learning model trained on a large corpus of clinical notes. By combining manually curated labels with NLP-derived labels, their model achieved a high AUROC for distant recurrence prediction, demonstrating the value of narrative text when manual chart review is infeasible.

    Current NLP approaches typically fall into two categories: rule-based systems and deep learning models (e.g., Transformers like BERT)~\cite{vaswani2017attention, devlin2019bert}. While deep learning offers high recall and generalizability, rule-based regular expression systems remain the gold standard for extracting standardized entities, such as TNM staging and receptor status, due to their deterministic precision and interpretability. For example, \citeA{yala2017using} demonstrated that for structured pathology parsing, domain-specific rule-based extractors could achieve near-perfect specificity. This precision is a critical requirement for generating reliable training labels in oncology, where misclassification of stage can severely bias risk models.
    
    Parallel to text mining, cancer registries provide curated data regarding diagnosis and staging but often suffer from reporting latency compared to real-time EHRs. Cross-institutional linkage studies, such as those by \citeA{charlton2022cancer}, have shown that combining registry and EHR data can yield high-performing models when labels are carefully harmonized. However, many of these approaches utilize existing datasets without explicitly detailing the upstream process of reconciling conflicting values or quantifying how multi-source extraction affects missingness patterns.

    \subsection{Multi-Source Clinical Data Integration}
    Recognizing the complementary nature of different data modalities, multi-source prediction models have emerged as a promising direction. Studies integrating clinical variables with imaging, radiomics, or molecular data have shown improved predictive performance over single-modality baselines. For example, \citeA{howard2023integration} fused routine clinical variables with deep learning features extracted from histology slides to predict recurrence scores, outperforming clinical-only nomograms. Additionally, \citeA{choi2025deep} reported that combining MRI-derived features with standard clinical data improved prediction accuracy in HER2-low breast cancer. More broadly, high-performing multi-modal architectures such as ICSDA~\cite{yao2022icsda} and recent deep-fusion recurrence models~\cite{zhang2025multimodal} demonstrate that pathological, clinical, imaging, and genomic streams can jointly improve risk prediction. A systematic review by \citeA{silveira2025harnessing} similarly concluded that multi-modal approaches generally achieve superior performance by capturing diverse biological signals.

    Despite these advances, the integration of heterogeneous data sources within routine EHR environments remains insufficiently studied. In particular, the reconciliation of discordant staging, biomarker, and treatment information across pathology reports, cancer registries, and clinical records poses unique methodological challenges. Unlike multi-omics or imaging-based fusion, where data streams are often additive, EHR-derived sources frequently overlap and exhibit source-dependent missingness. Existing models often operate on pre-curated feature sets or latent neural representations, which can obscure the provenance of individual clinical variables and limit auditability in healthcare settings~\cite{amann2020explainability}. Large registry-based datasets also contain substantial missingness in key prognostic fields, including stage and biomarker variables~\cite{yang2021prevalence}; therefore, models that rely only on structured variables may inherit systematic incompleteness. Similarly, prior EHR and registry integration systems improve data availability but often do not explicitly model how conflicting values should be reconciled across sources~\cite{gonzalez2023machine,goyal2024multi,linkov2018integration}. This study addresses these gaps by developing a robust, reproducible pipeline for harmonizing routine clinical data streams, with an emphasis on resolving cross-source inconsistencies, preserving feature provenance, and recovering missing prognostic variables through targeted text-based extraction. 
    
\section{Research Methodology} 
    \begin{figure*}[htb]
        \centering
        \includegraphics[width=0.6\textwidth]{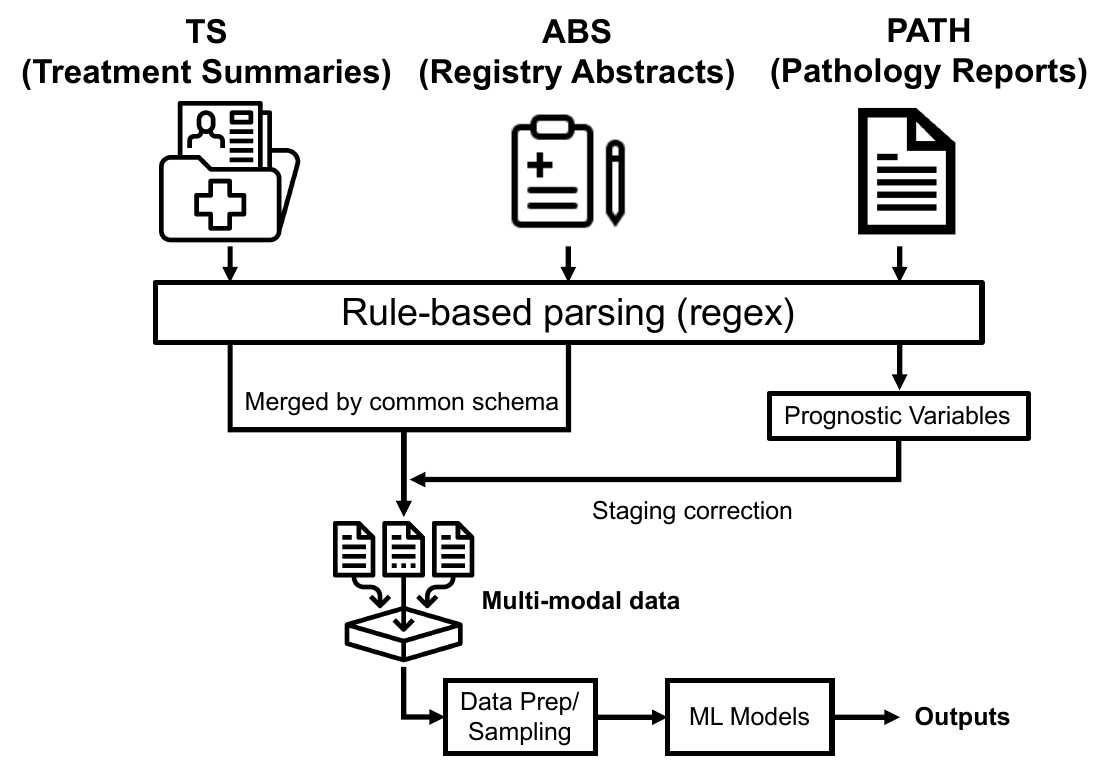}
        \caption{Pipeline overview with ingestion, reconciliation, preprocessing, and modeling. $\TS$, $\ABS$, and $\PATH$ are integrated using the precedence hierarchy \(PATH \succ ABS \succ TS\).}
        \label{fig:pipeline}
    \end{figure*}
    This section describes the proposed multi-modal data harmonization framework for automated breast cancer recurrence prediction. The unified multi-modal dataset is denoted as $\mathcal{D}=\{(\boldsymbol{x}^{(p)},y^{(p)})\}_{p=1}^{N_0}$, where $\boldsymbol{x}^{(p)}\in \mathbb{R}^D $ represents the $D$-dimensional vector of harmonized clinicopathologic features for patient $p$, and $y^{(p)}$ is the binary outcome label, with $y^{(p)}=1$ indicating a documented recurrence within 5 years post-surgery and $y^{(p)}=0$ otherwise. Fig.~\ref{fig:pipeline} illustrates the overall architecture of our framework, which consists of three core modules: Multi-Source Information Extraction, Precedence-Based Reconciliation, and Feature Construction.

    \subsection{Multi-Source Information Extraction}
    To capture a holistic view of the patient's clinical status, we integrate data from three heterogeneous sources: structured $\TS$, semi-structured $\ABS$, and unstructured $\PATH$. A tailored extraction mechanism is applied to each source to recover key prognostic variables, including TNM staging, histological grade, and biomarker status (ER, PR, and HER2) (see Table~\ref{tab:variables} in the Appendix for a complete description).  Figure~\ref{fig:extract_architecture} provides a detailed view of the modality-specific extraction architecture. The pipeline applies different deterministic logic to each source because the three modalities differ in structure, terminology, and error patterns. 
    
    \begin{figure*}[htb]
        \centering
        \includegraphics[width=0.95\textwidth]{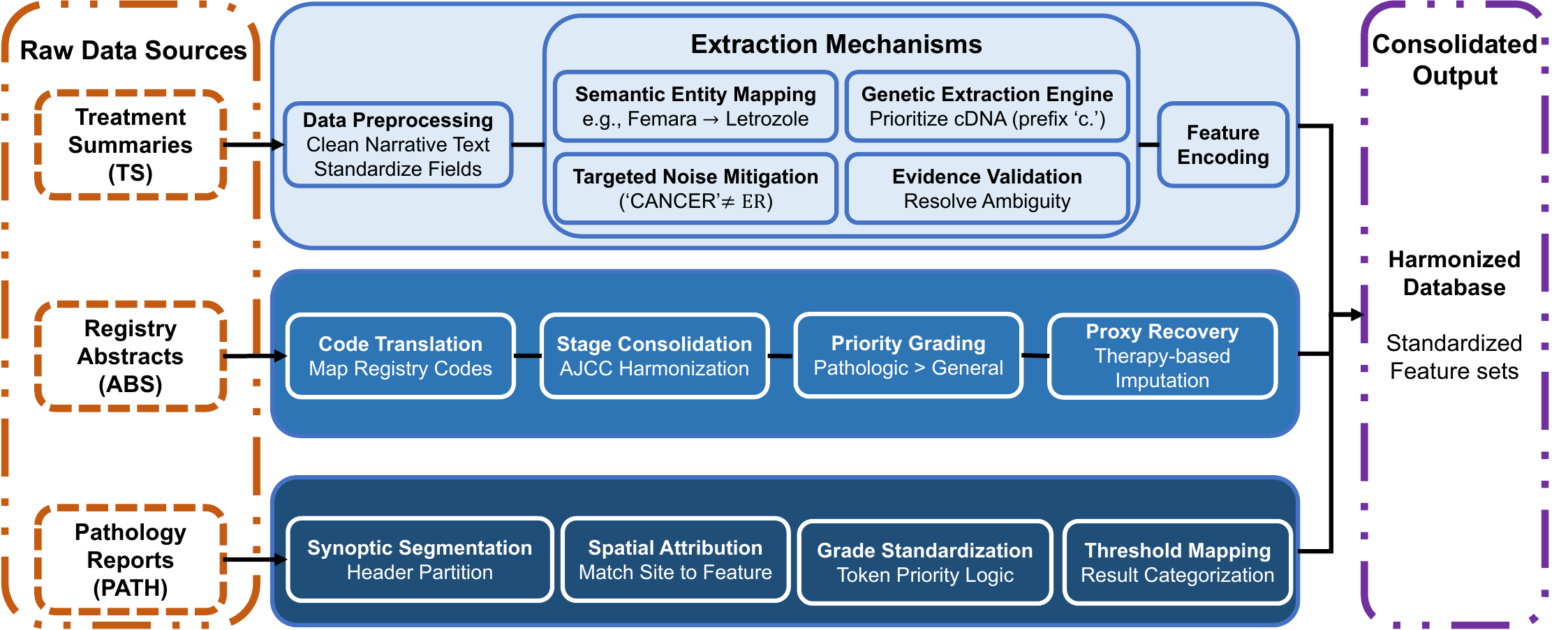}
        \caption{System architecture for multi-modality clinical data extraction and standardization. TS, ABS, and PATH are processed through source-specific extraction logic before being standardized and passed to the harmonized database.}
        \label{fig:extract_architecture}
    \end{figure*}

    \subsubsection{Structured and Semi-Structured Information Extraction}
    The $\TS$ and $\ABS$ provide a baseline of coded clinical data. Let $S_{\TS}$ and $S_{\ABS}$ denote the raw source repositories. We define extraction functions $g_{\TS}(\cdot)$ and $g_{\ABS}(\cdot)$ to parse these inputs into intermediate feature representations $T_{\TS}$ and $T_{\ABS}$, respectively.
    
    For the $\TS$ source, which follows a fixed-layout worksheet schema, we employ a deterministic, grid-based extraction procedure. Each patient record occupies a contiguous block of rows with a constant stride $G$ (e.g., $G = 46$). Feature values are extracted using predefined row-column offsets $(\Delta r_k, \Delta c_k)$ relative to the starting row index $s_p$ of patient $p$. Specifically, the value of the $k$-th feature is retrieved from the cell located at row $s_p + \Delta r_k$ and column $\Delta c_k$. This design enables reproducible and schema-aware extraction while minimizing ambiguity in feature localization. After values are localized from $S_{\TS}$ through the schema-aware extraction function $g_{\TS}(\cdot)$, the extracted entries are further standardized through terminology-guided normalization. This normalization is performed through a series of deterministic rules designed to maximize clinical fidelity. First, pharmaceutical trade names are mapped to their corresponding generic molecular entities; for example, Femara is normalized to Letrozole, consistent with the role of RxNorm in standardizing branded and generic drug names~\cite{rxnorm_overview}. Second, contextual negation and exclusion logic is applied to anatomical and pathological descriptors; for example, phrases such as ``no multifocal disease'' are explicitly identified as negated findings to prevent erroneous assignment of multifocal tumor characteristics, following the general logic of rule-based clinical negation detection~\cite{chapman2001simple}. Third, targeted lexical noise suppression is applied during biomarker parsing by filtering non-clinical tokens containing ambiguous substrings such as ``ER,'' including ``CANCER'' or ``CENTER,'' which could otherwise induce false-positive identification of ER status. Finally, the genetic extraction engine uses nomenclature-aware regular expressions to prioritize cDNA-level coordinates, such as BRCA2 c.6275\_6276delTT, over ambiguous narratives such as ``family history of BRCA,'' because Human Genome Variation Society (HGVS) nomenclature provides a standardized framework for describing sequence variants in clinical reports and databases~\cite{hart2024hgvs}.

    For the PDF-based $\ABS$ source, we apply a rigorous text normalization function $\Phi_{\text{pdf}}$ to address formatting inconsistencies inherent in document storage. This includes (i) repairing Optical Character Recognition (OCR) artifacts (e.g., ligature expansion), (ii) applying Unicode Normalization Form Compatibility Composition (NFKC), and (iii) de-hyphenating words broken across line breaks. Specifically, ligature expansion converts combined typographic glyphs into their constituent letters, such as converting an ``fi'' ligature into the two separate letters ``f'' and ``i,'' so that clinical keywords are not missed during rule-based matching. NFKC normalization is a Unicode text-standardization procedure that converts compatibility-equivalent characters into a consistent representation before keyword extraction and regular-expression matching. Following normalization, delimiter-bounded regular expressions are deployed to isolate fields under standardized registry headers, including pathology, surgery, treatment, and laboratory sections. To ensure consistency with the AJCC 8th edition guidelines, all extracted tokens are mapped to a standardized schema $\Sigma$. Specifically, $\Sigma$ acts as a controlled vocabulary where synonym variations are resolved to canonical forms; for instance, grade descriptors such as ``moderately differentiated'' or ``grade ii'' are mapped to $\{\mathrm{G2}\}$, and histology terms like ``infiltrating ductal'' are normalized to ``Invasive Ductal Carcinoma''.

    After documents from $S_{\ABS}$ are normalized by $\Phi_{\text{pdf}}$ and parsed through $g_{\ABS}(\cdot)$, the extraction logic further applies several source-specific standardization rules. First, registry-specific numeric treatment codes are translated into clinical descriptors; for example, treatment codes such as ``87'' are mapped to ``Patient Refused,'' whereas codes such as ``01'' are mapped to ``Therapy Administered'' across hormone and chemotherapy fields, following NAACCR treatment-coding definitions~\cite{naaccr_rx_hormone,naaccr_rx_chemo}. Second, granular TNM sub-stages, including $pT1mi$, $cT1a$, and $pT1b$, are consolidated into the broader $T1$ category for predictive modeling while preserving consistency with AJCC 8th edition breast cancer staging conventions~\cite{giuliano2018ajcc}. Third, histologic grade is recovered through a conditional fallback mechanism: the ``Pathologic Grade'' field is treated as the high-fidelity primary grade source, but if this field is unavailable or coded as $GX$, the general registry ``Grade'' field is used as a secondary source, consistent with NAACCR grade coding guidance and the preferred use of Nottingham grade for invasive breast tumors~\cite{naaccr_breast_grade_path}. Finally, missing HER2 status may be recovered from HER2-targeted therapy evidence only when the primary biomarker field is missing; for example, documented Trastuzumab or Herceptin treatment is treated as supporting evidence for HER2 positivity because trastuzumab is indicated for HER2-overexpressing breast cancer and requires HER2 testing for patient selection~\cite{herceptin_label_2024}. This proxy rule is applied exclusively to instances of missing biomarker data to reduce sparsity inherited from incomplete registry records and does not override directly reported pathology or registry biomarker values.

    \subsubsection{Unstructured Narrative Extraction from Pathology Reports}
    $\PATH$ serves as the high-fidelity ground truth for tumor characteristics but exists as unstructured free text. We treat each pathology report for patient $p$ as a document sequence $U^{(p)}$ and apply a multi-stage process. First, we apply a segmentation function $\Psi_{\text{path}}$ that partitions reports by standardized synoptic summary headers and anatomical or morphologic labels (e.g., ``SYNOPTIC SUMMARY LEFT BREAST'', ``Left Breast'', ``Right Breast'', or ``INVASIVE DUCTAL CARCINOMA'') to correctly associate findings with tumor laterality and tumor context. This anatomically aware segmentation prevents biomarker or grade conflation by ensuring that diagnostic features from distinct anatomical sites or morphologies are attributed to the correct tumor record.
    
    Second, a rule-based extraction module $\zeta_{\text{path}}$, utilizing domain-specific regular expressions, identifies mentions of TNM staging, histologic grade, and receptor status from these segmented regions. Within each segmented pathology region, the extraction module applies a clinically motivated priority hierarchy. For histologic grade, the highest priority is assigned to structured histologic or overall grade fields reported as Nottingham/Scarff--Bloom--Richardson grade or as an explicit Nottingham score, because the Nottingham combined histologic grade is based on tubule formation, nuclear pleomorphism, and mitotic count~\cite{cap_breast_invasive_protocol}. Explicit entries such as ``Overall Grade: Grade 3'' or ``Nottingham score 8/9'' are therefore treated as the most reliable grade signals. This priority rule is also consistent with Surveillance, Epidemiology, and End Results (SEER) coding guidance, which prioritizes Bloom--Richardson numeric scores over broader Bloom--Richardson grade labels when coding breast tumor grade~\cite{seer_breast_bloom_richardson}. If numeric grade fields are unavailable, equivalent grade notations, including Roman numerals such as Grade III, are mapped to the corresponding standardized category. As a final fallback, descriptive differentiation terms are used only when no explicit grade field is available; for example, ``well differentiated,'' ``moderately differentiated,'' and ``poorly differentiated'' are mapped to $\mathrm{G1}$, $\mathrm{G2}$, and $\mathrm{G3}$, respectively, following NAACCR grade coding conventions~\cite{naaccr_grade_manual}. This hierarchy prioritizes the most structured and site-specific grade evidence while reducing false assignments from less specific narrative descriptors. Biomarker results are standardized according to clinical reporting thresholds: ER and PR are mapped using ASCO/CAP receptor testing criteria, and HER2 immunohistochemistry (IHC) scores are categorized according to ASCO/CAP HER2 reporting guidance, where IHC 0/1+ is treated as negative, IHC 2+ is treated as equivocal requiring In Situ Hybridization (ISH) context, and IHC 3+ is treated as positive~\cite{allison2020estrogen,wolff2018human,fitzgibbons2014template}.

    The extraction process generates a structured vector for each pathology document:
    \begin{equation}
        \boldsymbol{v}_{\PATH}^{(p)} = \zeta_{\text{path}}(\Psi_{\text{path}}(U^{(p)}))
    \end{equation}
    where $\boldsymbol{v}_{\PATH}^{(p)}$ contains the extracted values. Crucially, domain-specific logic guards are applied to infer implicit values and enforce clinical consistency. For example, the system maps descriptive receptor intensity (e.g., ``Strong'' or ``Weak'') to binary ``Positive'' status. If the tumor ($T$) and nodal ($N$) stages are explicitly stated but metastasis ($M$) is absent, $M$ is inferred as $M0$ following standard reporting protocols. Furthermore, the parsing logic incorporates contextual exclusion rules to handle negation, ensuring that phrases such as ``No evidence of malignancy'' are not erroneously extracted as positive findings.     
  To reduce the risk of propagating extraction errors into downstream modeling, the extraction rules are refined through an iterative human-verification process. First, reports that cannot be parsed or are only partially parsed are reviewed to identify missing section headers, formatting variants, and syntactic patterns. Second, provisional extraction outputs are audited against the source documents to identify false-positive extractions (where unsupported values are incorrectly extracted) and false-negative extractions (where documented values are missed) with particular attention to negated findings, ambiguous receptor-status statements, multi-site pathology summaries, and laterality-specific descriptions. Third, the rule set is updated to address the identified error patterns and then reapplied to the audited reports. This review-and-update cycle continues until no additional extraction errors are identified in the audited cohort. This process is designed to reduce source-specific artifacts before the extracted variables enter the harmonized feature matrix.

    \subsection{Precedence-Based Conflict Reconciliation}
    A critical challenge in multi-source integration is the presence of discordant feature values across heterogeneous systems. To address this, we propose a hierarchical reconciliation strategy that prioritizes sources based on their proximity to the diagnostic event and information fidelity. We define a strict precedence order: $\PATH \succ \ABS \succ \TS$. This hierarchy reflects the clinical reality that $\PATH$ constitutes the diagnostic ground truth, while $\ABS$ and $\TS$ are derivative sources subject to transcription latency and aggregation errors. Because the reconciliation step may draw information from multiple documents for the same patient, we impose a temporal restriction to ensure that only baseline information is used for recurrence prediction. Specifically, the reconciliation process is anchored to the primary index surgery date, denoted by $T_0$, which separates diagnostic and index-treatment evidence from post-recurrence, metastatic, or subsequent-treatment records that could introduce future-data leakage.
    
    In scenarios involving multiple pathology documents for a single patient (e.g., an initial biopsy followed by a definitive surgical resection), our framework implements a pruning protocol that prioritizes reports containing definitive $T$ and $N$ staging. This effectively ensures that findings from surgical resection, the most reliable diagnostic event, supersede initial biopsy findings. A temporally restricted biomarker-filling logic is then applied: if primary receptor status is missing in the index surgical pathology report, the system retrieves receptor values only from baseline pathology documents or index-surgery pathology records anchored to $T_0$, such as a pre-surgical biopsy or the definitive surgical pathology report. This rule improves completeness while preventing recurrence-biopsy, metastasis-evaluation, or subsequent-treatment information from entering the prediction features. The rationale for this baseline-only filling rule is that primary breast cancer receptor markers, including ER, PR, and HER2, are generally expected to remain concordant between diagnostic biopsy and definitive surgical pathology when no intervening neoadjuvant systemic therapy has occurred~\cite{dekker2013reliability,asogan2017concordance}. Therefore, cross-document filling is restricted to records representing the same baseline disease episode and is not applied using post-recurrence biopsies, metastatic evaluations, or records after intervening systemic therapy, since such records may reflect treatment-induced or progression-related biomarker changes.

    For a given patient $p$ and feature $j$ (e.g., ER status), the harmonized value $x_j^{(p)}$ is derived using a hierarchical (waterfall) logic: 
    \begin{equation}
        x_j^{(p)} = 
        \begin{cases} 
        v_{\PATH, j}^{(p)} & \text{if } v_{\PATH, j}^{(p)} \notin \mathcal{N} \\
        v_{\ABS, j}^{(p)} & \text{if } v_{\PATH, j}^{(p)} \in \mathcal{N} \land v_{\ABS, j}^{(p)} \notin \mathcal{N} \\
        v_{\TS, j}^{(p)} & \text{otherwise}
        \end{cases}
    \end{equation}
    where $v_{\PATH,j}^{(p)}$, $v_{\ABS,j}^{(p)}$, and $v_{\TS,j}^{(p)}$ denote the source-specific candidate values for feature $j$ after extraction and standardization, and $\mathcal{N}$ represents the set of non-informative values (e.g., null, ``Not Stated'', ``Unknown'', ``NX''). This logic ensures that definitive findings from the primary pathology report supersede potentially outdated or transposed values found in downstream systems. The reconciliation process results in a unified feature vector $\boldsymbol{x}^{(p)}$ with maximized information density. To examine whether the reconciliation hierarchy materially affects downstream prediction, we conducted a sensitivity analysis using an alternative fallback order, $\PATH \succ \TS \succ \ABS$, while keeping $\PATH$ as the highest-priority diagnostic source. The detailed results are reported in Appendix~\ref{app:precedence_sensitivity}.    
    
    \subsection{Feature Construction and Preprocessing}
    Following the precedence-based reconciliation, the unified dataset $\mathcal{D}$ consolidates the highest-fidelity information available for each patient. However, residual missingness persists in cases where a feature is absent across all three source systems. To address this challenge, we apply type-specific imputation strategies to the reconciled data. Continuous variables, such as tumor size and patient age, are handled using mean imputation to preserve the central tendency of the distribution. For biomarker percentages or ratios, we use status-conditional imputation only when the categorical biomarker status is observed but the corresponding numeric value is missing. For example, if HER2 status is available but the HER2 ratio is missing, the missing numeric value is imputed using the mean value among training-set patients with the same HER2 status. This approach is used because biomarker measurements are biologically and clinically linked to their categorical status, and global mean imputation will incorrectly pool biologically distinct positive and negative cases. The procedure is applied only to numeric biomarker fields and does not infer, overwrite, or relabel the categorical receptor status itself. Importantly, all imputation parameters are estimated from the training data only and are then applied unchanged to validation or test data to avoid information leakage.
    
    To transform clinical attributes into a format suitable for machine learning, we employ a dual-encoding strategy contingent on variable ordinality. Variables characterized by intrinsic hierarchy, specifically tumor grade ($\mathrm{Grade\ 1} < \mathrm{Grade\ 2} < \mathrm{Grade\ 3}$) and the $T$ and $N$ staging, are mapped to integer sequences via ordinal encoding. This approach preserves the monotonic relationship between disease severity and the recurrence outcome. In contrast, nominal variables lacking natural ordering, such as histology type, laterality, and surgical information, are processed using one-hot encoding to avoid introducing spurious ordinal correlations. Subsequently, all continuous features undergo Z-score normalization~\cite{cabello2023impact} to standardize their range (mean 0, variance 1), thereby mitigating the risk of high-magnitude features disproportionately influencing the optimization landscape. 
    
\subsection{Recurrence Classification Framework}
    To predict the likelihood of breast cancer recurrence, we formulate the problem as a supervised binary classification task. Let $f(\cdot; \theta)$ denote a classification model parameterized by $\theta$. For each patient $p$ with a harmonized feature vector $\boldsymbol{x}^{(p)}$, the model outputs a continuous recurrence risk score $\hat{y}^{(p)} \in [0, 1]$:
    \begin{equation}
        \hat{y}^{(p)} = f(\boldsymbol{x}^{(p)}; \theta)
    \end{equation}
    where higher values of $\hat{y}^{(p)}$ indicate greater model-estimated recurrence risk within the 5-year post-surgery window.

    To optimize the model parameters $\theta$, particularly for gradient-based learners (e.g., TabNet, XGBoost (XGB)), we minimize the Binary Cross-Entropy (Log-Loss) function, which penalizes disagreement between the model output scores and the true binary labels:
    \begin{equation}
        \mathcal{L}(\theta) = - \frac{1}{N_0} \sum_{i=1}^{N_0} \left[ y^{(i)} \log(\hat{y}^{(i)}) + (1 - y^{(i)}) \log(1 - \hat{y}^{(i)}) \right]
    \end{equation}
    Finally, to translate the continuous risk score into a binary decision $\hat{c}^{(p)}$ when threshold-dependent metrics are reported, we apply a decision threshold $\tau$:
    \begin{equation}
        \hat{c}^{(p)} = \mathbb{I}(\hat{y}^{(p)} > \tau),
    \end{equation}    
    where $\mathbb{I}(\cdot)$ is the indicator function. It is important to note that, although optimization with binary cross-entropy yields outputs on a probability scale, it does not inherently guarantee probabilistic calibration. This distinction is clinically important because discrimination and calibration address different aspects of prediction performance: a model may rank patients accurately while still overestimating or underestimating absolute recurrence probabilities. Therefore, before the framework can be used for threshold-dependent clinical decision-making or individualized absolute-risk estimation, future work should include formal calibration assessment, such as calibration curves, Brier score evaluation, and potential post-hoc recalibration methods~\cite{steyerberg2010assessing,van2019calibration,guo2017calibration}.

\section{Materials and Experimental Design}
\label{sec:materials}
    In this study, we use retrospective clinical data from UTMC, encompassing a diverse cohort of patients diagnosed with invasive breast cancer between 2007 and 2025. This repository integrates three distinct clinical streams: structured $\TS$, annual $\ABS$, and raw $\PATH$. The depth and breadth of this multi-source information provide a robust platform for designing data-driven strategies for recurrence prediction. Identification of recurrence cases within this dataset is based on a 5-year post-surgery follow-up window, with a designation of ``0'' for patients remaining disease-free and ``1'' for those with documented locoregional or distant recurrence. Consistent with the reconciliation framework, $T_0$ denotes the primary index surgery date. All predictive variables are restricted to baseline clinical information from the diagnostic workup, index surgical pathology, and initial treatment course anchored to $T_0$. Patients are assigned to the non-recurrence class only if they have confirmed recurrence-free follow-up exceeding five years, while patients with less than five years of recurrence-free follow-up are treated as right-censored and excluded from the analytic cohort.
    
    By integrating clinical input with findings from prior studies~\cite{alzu2021predicting, gonzalez2023machine}, we selected a comprehensive set of variables for our study, as shown in Table \ref{tab:variables} in the Appendix, including AJCC 8th edition staging components, tumor grade, receptor status (ER, PR, and HER2), and treatment descriptors. The final analytic dataset comprises a total of $N=6,060$ cases (representing 5,845 unique patients), among whom a minority are identified as recurrence-positive ($n=364$, 6.0\%), while the majority are non-recurrence ($n=5,696$, 94.0\%). To rigorously evaluate model performance, we adopted a two-stage partitioning strategy. First, for hyperparameter optimization, the dataset was randomly partitioned into training (70\%), validation (15\%), and testing (15\%) subsets. This split was utilized strictly for Bayesian optimization to identify the optimal model configurations without biasing the final evaluation. Following parameter selection, the dataset was re-partitioned into a final training (80\%) and independent testing (20\%) split to assess the generalized performance of the optimized models.

    To systematically assess model robustness under varying degrees of class imbalance, we implement a controlled undersampling strategy across the training, validation, and testing partitions. We construct three distinct experimental scenarios with specific minority-to-majority ratios: 1:1 (balanced), 1:2, and 1:3. For each scenario, the majority class (non-recurrence) is randomly downsampled in all sets to match the target ratio, while the minority class (recurrence) is kept constant. This design ensures that performance metrics are evaluated under consistent distribution constraints, allowing us to isolate the impact of data fusion from the confounding effects of varying prevalence. For the controlled 1:1, 1:2, and 1:3 class-imbalance settings, model performance is evaluated using AUROC, AUPRC, F$_1$ score, and G-Mean. In addition, we assess performance on the original imbalanced cohort without undersampling. For this setting, we primarily emphasize threshold-independent metrics, namely AUROC and AUPRC, because they evaluate discriminative performance across all possible decision thresholds without requiring the selection of a specific binary operating point. This consideration is particularly important in highly imbalanced clinical prediction tasks, where threshold-dependent metrics may be sensitive to class prevalence and operating-point selection.

    We compare the predictive performance of four machine learning methods: RF, XGB, Regularized Greedy Forests (RGF), and TabNet~\cite{breiman2001random, chen2016xgboost, johnson2013learning, arik2021tabnet}. To quantify the value of data harmonization, each model is evaluated across three distinct feature configurations: single-source, which utilizes only the features from the structured TS; multi-source, which incorporates all available features following the data fusion pipeline; and benchmark, a restricted subset comprising only those features commonly reported in prior breast cancer studies. Specifically, the benchmark set includes 16 variables identified from key literature~\cite{alzu2021predicting, gonzalez2023machine}. These variables are highlighted with an asterisk (*) in Table~\ref{tab:variables} in the Appendix. Hyperparameters are tuned using the validation set via Bayesian optimization~\cite{frazier2018bayesian,shao2025multi,xie2022physics}. All classification experiments are carried out on a standard workstation using Python as the programming language, with random seeds fixed across feature sets to ensure a fair comparison. Finally, in addition to the primary hold-out analysis, we perform 5-fold cross-validation to further corroborate the robustness of our findings.
    
\section{Results and Discussion}
    \subsection{Missingness Reduction}
    \begin{figure*}[tb]
        \centering
        \includegraphics[width=0.8\textwidth]{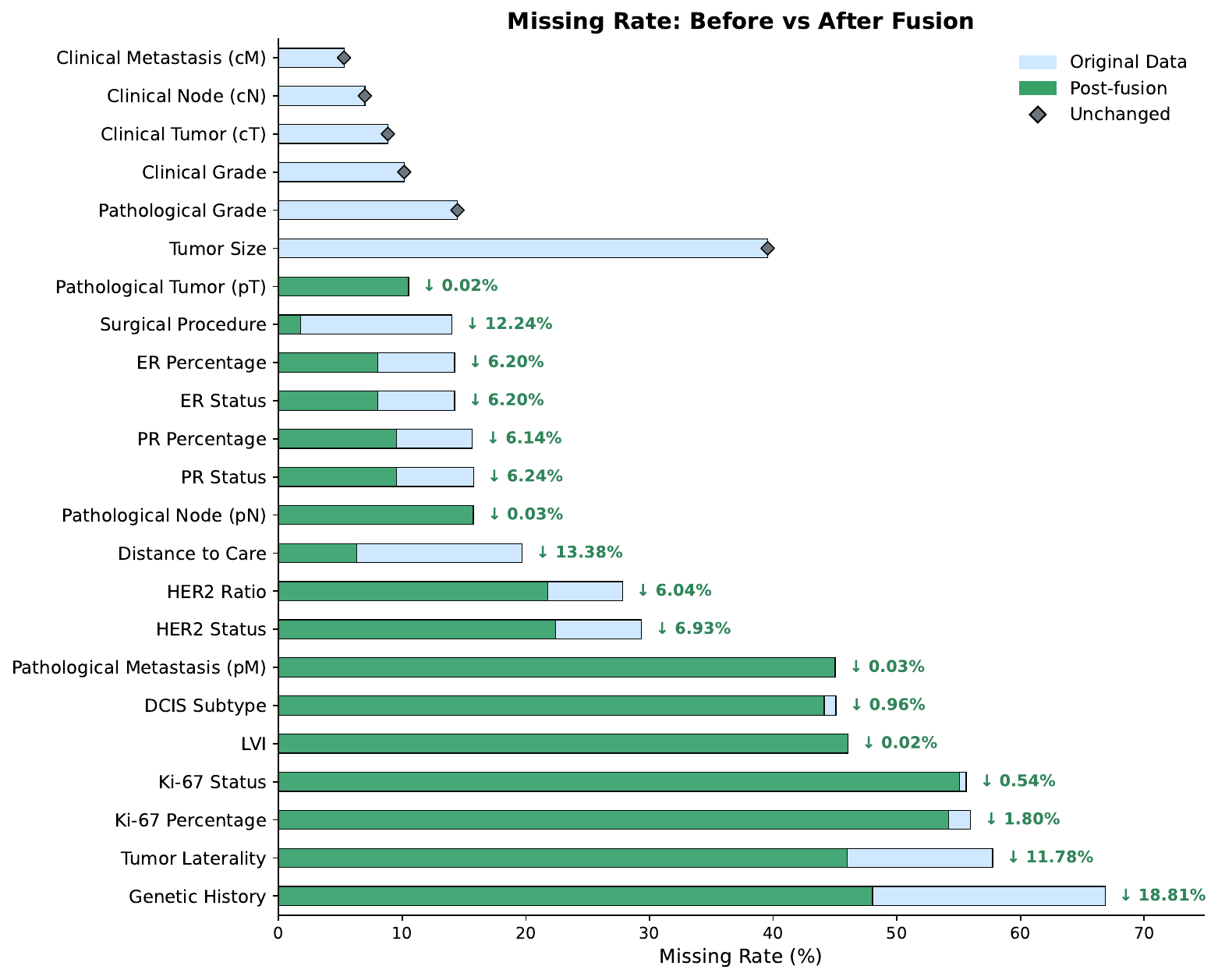}
        \caption{Feature-wise missing rate before (blue) and after fusion (green). Gray diamonds denote unchanged features. Labels ``$\downarrow$ X.XX\%'' give the absolute reduction.}
        \label{fig:missingness}
    \end{figure*}
    
    Figure~\ref{fig:missingness} demonstrates how multi-modal data fusion reduces missingness across key clinical features (see Table~\ref{tab:variables} for variable definitions). Relative to the original single-source dataset, the fused multi-source dataset shows substantial absolute reductions in missing values for core biomarkers: ER status and ER percentage decrease by 6.20\%, PR status by 6.24\%, and PR percentage by 6.14\%. HER2 fields also improve, with HER2 status decreasing by 6.93\% and HER2 ratio by 6.04\%. Beyond biomarkers, we observe the largest reductions for genetic history (18.81\%), distance to care (13.38\%), surgical procedure (12.24\%), and tumor laterality (11.78\%). More modest decreases occur for Ki-67 percentage (1.80\%), Ductal carcinoma in situ (DCIS) subtype (0.96\%), and Ki-67 status (0.54\%), with marginal gains ($<0.1\%$) for pathological metastasis (pM), LVI, pathological node (pN), and pathological tumor (pT). Several variables remain unchanged after fusion, including clinical tumor (cT), clinical node (cN), clinical metastasis (cM), clinical grade, pathological grade, and tumor size, indicating that additional sources did not alter the completeness of these fields. Overall, multi-source fusion increases completeness for high-impact biomarkers and related clinical fields, reduces reliance on imputation, and supports more robust downstream modeling; unchanged fields highlight targets for future curation or extraction. These completeness gains increase the effective sample size available for training, which we next show translates into improved discrimination.
    
    \subsection{Predictive Performance}
    We report the predictive performance yielded by four models (RF, XGB, RGF, and TabNet) across the controlled imbalance scenarios and the original imbalanced cohort. To isolate the incremental value of feature fusion from sampling variability, the random seeds and sampled indices are kept identical across all feature configurations, multi-source, single-source, and benchmark, within a split. Consistent with our controlled experimental design, undersampling is applied across the training, validation, and testing partitions to strictly enforce these target ratios. 
        
    Table~\ref{tab:main_results} reports performance on the hold-out test set, which is the 20 percent split reserved for final evaluation. Across models and imbalance settings, multi-source generally outperforms single-source in discrimination. Averaged over all 12 model and ratio combinations, multi-source provides absolute gains of about 0.039 in AUROC and 0.056 in AUPRC relative to single-source. These aggregate improvements are consistent with the per-model performance differences shown in Table~\ref{tab:delta_holdout}, which reports the differences between multi-source and single-source models.
    
    At a ratio of 1:1, RF improves from an AUROC of 0.790 and an AUPRC of 0.754 (single-source) to 0.879 and 0.891 (multi-source), and TabNet improves from 0.795 and 0.797 to 0.892 and 0.898. XGB increases from 0.792 and 0.760 to 0.865 and 0.861, and RGF from 0.773 and 0.753 to 0.866 and 0.884. At a ratio of 1:2, XGB rises from 0.822 and 0.718 to 0.851 and 0.768; RF from 0.836 and 0.741 to 0.851 and 0.768; TabNet from 0.812 and 0.703 to 0.825 and 0.726. RGF is essentially tied on AUROC (0.837 single-source vs 0.836 multi-source) but is higher for multi-source on AUPRC, F$_1$, and G-Mean (0.748, 0.603, 0.691 vs 0.758, 0.632, 0.713). At a ratio of 1:3, TabNet increases from 0.802 and 0.642 to 0.831 and 0.685, and XGB from 0.845 and 0.649 to 0.854 and 0.685. RF shows a smaller discrimination gain at 1:3 (AUPRC 0.653 to 0.657), and RGF is flat on AUROC (0.838 to 0.838) but improves on AUPRC (0.652 to 0.663). Two exceptions appear under the most imbalanced setting (1:3), where threshold-dependent point metrics can be sensitive to prevalence. For the ratio 1:3, RF achieves a slightly higher AUPRC with the benchmark features (comprising 16 standard clinical variables) than with multi-source (0.669 vs 0.657), and RGF attains its best F$_1$ and G-Mean on the benchmark features (0.585 and 0.679). Otherwise, the benchmark typically trails both single-source and multi-source. Overall, these hold-out results indicate that feature fusion yields the most consistent advantages for ranking performance, with large improvements at a ratio of 1:1 and smaller but generally positive gains as imbalance increases. 
    
    Under the original imbalanced cohort without undersampling, the multi-source configuration also achieves the strongest threshold-independent performance across all evaluated algorithms. The multi-source RF model obtains the highest AUROC of 0.845, while the multi-source RGF model obtains the highest AUPRC of 0.314. Because the recurrence prevalence is approximately 6\%, the random AUPRC baseline is $\sim$0.06; therefore, the observed AUPRC values indicate substantially improved rare-event retrieval relative to random ranking. These results suggest that the performance gains from multi-source harmonization are not an artifact of controlled undersampling, but persist under the natural clinical class distribution.
    
    \begin{table*}[tb]
    \centering
    \caption{Comparison of model performance (AUROC, AUPRC, F$_1$, G-Mean) under different class imbalance ratios. Multi-source models use all available features after data fusion; single-source models use only original features; benchmark models use only commonly reported features. 
    }  
    \label{tab:main_results}
    \resizebox{\textwidth}{!}{
    \begin{tabular}{@{} c | c | cccc | cccc | cccc @{}}
    \toprule
    \multirow{2}{*}{Imbalance}
    & \multirow{2}{*}{Model}
    & \multicolumn{4}{c}{Multi-source}
    & \multicolumn{4}{c}{Single-source}
    & \multicolumn{4}{c}{Benchmark} \\
    \cmidrule(lr){3-6} \cmidrule(lr){7-10} \cmidrule(lr){11-14}
    & & AUROC & AUPRC & F$_1$ & G-Mean & AUROC & AUPRC & F$_1$ & G-Mean & AUROC & AUPRC & F$_1$ & G-Mean \\
    \midrule
    
    \multirow{4}{*}{1:1}
    & RF     & \textbf{0.879} & \textbf{0.891} & \textbf{0.786} & \textbf{0.788} & 0.790 & 0.754 & 0.743 & 0.740 & 0.759 & 0.772 & 0.679 & 0.675 \\
    & XGB    & \textbf{0.865} & \textbf{0.861} & \textbf{0.778} & \textbf{0.781} & 0.792 & 0.760 & 0.768 & 0.760 & 0.772 & 0.760 & 0.662 & 0.662 \\
    & RGF    & \textbf{0.866} & \textbf{0.884} & \textbf{0.726} & \textbf{0.743} & 0.773 & 0.753 & 0.711 & 0.697 & 0.767 & 0.749 & 0.711 & 0.699 \\
    & TabNet & \textbf{0.892} & \textbf{0.898} & \textbf{0.839} & \textbf{0.842} & 0.795 & 0.797 & 0.725 & 0.738 & 0.725 & 0.731 & 0.654 & 0.662 \\
    
    \midrule
    \multirow{4}{*}{1:2}
    & RF     & \textbf{0.851} & \textbf{0.768} & \textbf{0.688} & \textbf{0.767} & 0.836 & 0.741 & 0.653 & 0.737 & 0.823 & 0.761 & 0.648 & 0.730 \\
    & XGB    & \textbf{0.851} & \textbf{0.768} & \textbf{0.688} & \textbf{0.767} & 0.822 & 0.718 & 0.670 & 0.751 & 0.805 & 0.697 & 0.615 & 0.700 \\
    & RGF    & \textbf{0.836} & \textbf{0.758} & \textbf{0.632} & \textbf{0.713} & \textbf{0.837} & 0.748 & 0.603 & 0.691 & 0.825 & 0.739 & 0.615 & 0.696 \\
    & TabNet & \textbf{0.825} & \textbf{0.726} & \textbf{0.629} & \textbf{0.715} & 0.812 & 0.703 & 0.614 & 0.692 & 0.773 & 0.698 & 0.599 & 0.688 \\
    
    \midrule
    \multirow{4}{*}{1:3}
    & RF     & \textbf{0.849} & 0.657 & \textbf{0.659} & \textbf{0.784} & 0.830 & 0.653 & 0.588 & 0.675 & 0.815 & \textbf{0.669} & 0.608 & 0.739 \\
    & XGB    & \textbf{0.854} & \textbf{0.685} & \textbf{0.609} & \textbf{0.704} & 0.845 & 0.649 & 0.587 & 0.685 & 0.829 & 0.672 & 0.583 & 0.673 \\
    & RGF    & \textbf{0.838} & \textbf{0.663} & 0.571 & 0.664 & \textbf{0.838} & 0.652 & 0.534 & 0.634 & 0.828 & 0.656 & \textbf{0.585} & \textbf{0.679} \\
    & TabNet & \textbf{0.831} & \textbf{0.685} & \textbf{0.613} & \textbf{0.719} & 0.802 & 0.642 & 0.522 & 0.623 & 0.793 & 0.676 & 0.569 & 0.657 \\

    \midrule
    \multirow{4}{*}{Original}
    & RF     & \textbf{0.845} & \textbf{0.297} & -- & -- & 0.827 & 0.287 & -- & -- & 0.760 & 0.243 & -- & -- \\
    & XGB    & \textbf{0.825} & \textbf{0.311} & -- & -- & 0.771 & 0.271 & -- & -- & 0.751 & 0.229 & -- & -- \\
    & RGF    & \textbf{0.828} & \textbf{0.314} & -- & -- & 0.824 & 0.303 & -- & -- & 0.796 & 0.300 & -- & -- \\
    & TabNet & \textbf{0.825} & \textbf{0.281} & -- & -- & 0.803 & 0.268 & -- & -- & 0.778 & 0.258 & -- & -- \\

    \bottomrule
    \end{tabular}
    }
    \end{table*}

    \begin{table}[tb]
    \centering
    \caption{The absolute performance differences of the Hold-out test set.}
    \label{tab:delta_holdout}
    \begin{threeparttable}
        \scriptsize
        \setlength{\tabcolsep}{3pt}
        \renewcommand{\arraystretch}{0.88}
    
        \begin{tabular*}{0.48\textwidth}{@{\extracolsep{\fill}} l l c c c c @{}} 
            \toprule
            Imbalance & Model & $\Delta$AUROC & $\Delta$AUPRC & $\Delta$F$_1$ & $\Delta$G-Mean \\
            \midrule
            \multirow{4}{*}{1:1} 
                   & RF     & +0.089 & +0.137 & +0.043 & +0.048 \\
                   & XGB    & +0.073 & +0.101 & +0.010 & +0.021 \\
                   & RGF    & +0.093 & +0.131 & +0.015 & +0.046 \\
                   & TabNet & +0.097 & +0.101 & +0.114 & +0.104 \\
            \midrule
            \multirow{4}{*}{1:2} 
                   & RF     & +0.015 & +0.027 & +0.035 & +0.030 \\
                   & XGB    & +0.029 & +0.050 & +0.018 & +0.016 \\
                   & RGF    & -0.001 & +0.010 & +0.029 & +0.022 \\
                   & TabNet & +0.013 & +0.023 & +0.015 & +0.023 \\
            \midrule
            \multirow{4}{*}{1:3} 
                   & RF     & +0.019 & +0.004 & +0.071 & +0.109 \\
                   & XGB    & +0.009 & +0.036 & +0.022 & +0.019 \\
                   & RGF    & +0.000 & +0.011 & +0.037 & +0.030 \\
                   & TabNet & +0.029 & +0.043 & +0.091 & +0.096 \\
    
            \midrule
            \multirow{4}{*}{Original}
                   & RF     & +0.018 & +0.010 & -- & -- \\
                   & XGB    & +0.054 & +0.040 & -- & -- \\
                   & RGF    & +0.004 & +0.011 & -- & -- \\
                   & TabNet & +0.022 & +0.013 & -- & -- \\
            \bottomrule
        \end{tabular*}
    
        \begin{tablenotes}[flushleft]\scriptsize
            \item \textit{Note:} $\Delta$ is computed as multi-source minus single-source. Values are rounded to three decimals.
            \item \textbf{Overall mean gains:} $\Delta$AUROC $\approx$ +0.039, $\Delta$AUPRC $\approx$ +0.056, $\Delta$F$_1$ $\approx$ +0.042, $\Delta$G-Mean $\approx$ +0.047. The original imbalanced cohort is reported separately because F$_1$ and G-Mean are not emphasized for this setting.
        \end{tablenotes}
    \end{threeparttable}
    \end{table}

    Five-fold cross-validation (Table~\ref{tab:cv_all}) corroborates the hold-out findings. Across all models and imbalance settings, multi-source exceeds single-source in both AUROC and AUPRC in all 12 comparisons. The AUROC gains range from 0.010 to 0.042, and the AUPRC gains range from 0.006 to 0.034. For illustration: at a ratio of 1:1, TabNet moves from AUROC $0.727\pm0.038$ and AUPRC $0.721\pm0.047$ (single-source) to $0.769\pm0.031$ and $0.755\pm0.025$ (multi-source), and RF moves from $0.803\pm0.023$ and $0.794\pm0.021$ to $0.821\pm0.031$ and $0.803\pm0.018$. At a ratio of 1:2, RGF moves from $0.804\pm0.022$ and $0.694\pm0.036$ to $0.814\pm0.021$ and $0.717\pm0.031$, and XGB from $0.793\pm0.017$ and $0.676\pm0.025$ to $0.808\pm0.022$ and $0.702\pm0.030$. At a ratio of 1:3, XGB reaches $0.816\pm0.016$ and $0.632\pm0.037$ (multi-source) versus $0.801\pm0.011$ and $0.622\pm0.019$ (single-source), and TabNet increases from $0.735\pm0.023$ and $0.506\pm0.050$ to $0.777\pm0.040$ and $0.533\pm0.053$. 
    
    Additionally, threshold-dependent metrics show smaller but consistent improvements. G-Mean is higher for multi-source in all 12 settings; for example, at a ratio of 1:1, it increases from $0.668$ to $0.710$ for TabNet and from $0.695$ to $0.734$ for RGF. F$_1$ improves in 11 of 12 settings and is essentially unchanged for RF at a ratio of 1:3 ($0.513$ single-source vs $0.512$ multi-source). As the imbalance worsens from 1:1 to 1:3, mean performance declines for all feature sets; however, the relative advantage of fusion remains, most clearly in AUPRC, which is the metric most sensitive to minority prevalence. Fold standard deviations for AUROC and AUPRC are generally modest (about $0.008$ to $0.047$), with larger variability for F$_1$ in the most imbalanced cases (up to $0.100$). The multi-source gains in AUROC and AUPRC commonly exceed $0.01$ to $0.02$, indicating stable improvements across folds without reliance on a single operating threshold. These fold-level advantages indicate that improvements are not attributable to a single split or a single operating threshold but persist under resampling variability.
    
    \begin{table*}[tb]
    \centering
    \caption{Five-fold cross-validation across class-imbalance settings (1:1, 1:2, 1:3). Random undersampling is applied within CV folds; threshold-dependent metrics are evaluated using the standard classification threshold $\tau=0.5$. }
    \label{tab:cv_all}
    \resizebox{\textwidth}{!}{%
    \begin{tabular}{@{} l | l | cccc | cccc | cccc @{}}
    \toprule
    \multirow{2}{*}{Ratio} & \multirow{2}{*}{Model} &
    \multicolumn{4}{c}{Multi-source} &
    \multicolumn{4}{c}{Single-source} &
    \multicolumn{4}{c}{Benchmark} \\
    \cmidrule(lr){3-6}\cmidrule(lr){7-10}\cmidrule(lr){11-14}
    & & AUROC & AUPRC & F$_1$ & G-Mean & AUROC & AUPRC & F$_1$ & G-Mean & AUROC & AUPRC & F$_1$ & G-Mean \\
    \midrule
    
    \multirow{4}{*}{1:1}
    & RF
    & \textbf{\ms{0.821}{0.03}} & \textbf{\ms{0.803}{0.02}} & \textbf{\ms{0.760}{0.04}} & \textbf{\ms{0.756}{0.04}}
    & \ms{0.803}{0.02} & \ms{0.794}{0.02} & \ms{0.741}{0.04} & \ms{0.743}{0.03}
    & \ms{0.766}{0.05} & \ms{0.741}{0.06} & \ms{0.702}{0.04} & \ms{0.707}{0.04} \\ \addlinespace[3pt]
    & XGB
    & \textbf{\ms{0.806}{0.04}} & \textbf{\ms{0.796}{0.04}} & \textbf{\ms{0.728}{0.03}} & \textbf{\ms{0.729}{0.03}}
    & \ms{0.790}{0.03} & \ms{0.786}{0.03} & \ms{0.717}{0.04} & \ms{0.715}{0.04}
    & \ms{0.751}{0.06} & \ms{0.722}{0.07} & \ms{0.692}{0.04} & \ms{0.695}{0.04} \\ \addlinespace[3pt]
    & RGF
    & \textbf{\ms{0.789}{0.05}} & \textbf{\ms{0.764}{0.05}} & \textbf{\ms{0.730}{0.06}} & \textbf{\ms{0.734}{0.06}}
    & \ms{0.756}{0.03} & \ms{0.740}{0.04} & \ms{0.696}{0.04} & \ms{0.695}{0.04}
    & \ms{0.733}{0.05} & \ms{0.707}{0.06} & \ms{0.672}{0.06} & \ms{0.673}{0.06} \\ \addlinespace[3pt]
    & TabNet
    & \textbf{\ms{0.769}{0.03}} & \textbf{\ms{0.755}{0.03}} & \textbf{\ms{0.707}{0.06}} & \textbf{\ms{0.710}{0.04}}
    & \ms{0.727}{0.04} & \ms{0.721}{0.05} & \ms{0.657}{0.03} & \ms{0.668}{0.02}
    & \ms{0.707}{0.05} & \ms{0.706}{0.06} & \ms{0.637}{0.06} & \ms{0.637}{0.05} \\
    \midrule
    
    \multirow{4}{*}{1:2}
    & RF
    & \textbf{\ms{0.825}{0.03}} & \textbf{\ms{0.722}{0.04}} & \textbf{\ms{0.611}{0.05}} & \textbf{\ms{0.691}{0.04}}
    & \ms{0.815}{0.03} & \ms{0.716}{0.04} & \ms{0.591}{0.06} & \ms{0.672}{0.05}
    & \ms{0.783}{0.04} & \ms{0.665}{0.04} & \ms{0.561}{0.03} & \ms{0.652}{0.02} \\ \addlinespace[3pt]
    & XGB
    & \textbf{\ms{0.808}{0.02}} & \textbf{\ms{0.702}{0.03}} & \textbf{\ms{0.631}{0.04}} & \textbf{\ms{0.719}{0.03}}
    & \ms{0.793}{0.02} & \ms{0.676}{0.03} & \ms{0.626}{0.02} & \ms{0.713}{0.02}
    & \ms{0.775}{0.02} & \ms{0.650}{0.03} & \ms{0.601}{0.03} & \ms{0.685}{0.02} \\ \addlinespace[3pt]
    & RGF
    & \textbf{\ms{0.814}{0.02}} & \textbf{\ms{0.717}{0.03}} & \textbf{\ms{0.617}{0.06}} & \textbf{\ms{0.699}{0.05}}
    & \ms{0.804}{0.02} & \ms{0.694}{0.04} & \ms{0.604}{0.05} & \ms{0.689}{0.04}
    & \ms{0.788}{0.03} & \ms{0.690}{0.04} & \ms{0.563}{0.05} & \ms{0.655}{0.04} \\ \addlinespace[3pt]
    & TabNet
    & \textbf{\ms{0.779}{0.03}} & \textbf{\ms{0.635}{0.03}} & \textbf{\ms{0.572}{0.08}} & \textbf{\ms{0.660}{0.07}}
    & \ms{0.748}{0.05} & \ms{0.623}{0.06} & \ms{0.527}{0.10} & \ms{0.626}{0.09}
    & \ms{0.725}{0.03} & \ms{0.599}{0.03} & \ms{0.501}{0.07} & \ms{0.605}{0.07} \\
    \midrule
    
    \multirow{4}{*}{1:3}
    & RF
    & \textbf{\ms{0.825}{0.01}} & \textbf{\ms{0.654}{0.02}} & \ms{0.512}{0.03} & \textbf{\ms{0.617}{0.03}}
    & \ms{0.813}{0.02} & \ms{0.639}{0.02} & \textbf{\ms{0.513}{0.01}} & \ms{0.616}{0.01}
    & \ms{0.790}{0.01} & \ms{0.601}{0.04} & \ms{0.482}{0.02} & \ms{0.594}{0.02} \\ \addlinespace[3pt]
    & XGB
    & \textbf{\ms{0.816}{0.02}} & \textbf{\ms{0.632}{0.04}} & \textbf{\ms{0.557}{0.03}} & \textbf{\ms{0.665}{0.03}}
    & \ms{0.801}{0.01} & \ms{0.622}{0.02} & \ms{0.554}{0.02} & \ms{0.660}{0.02}
    & \ms{0.784}{0.01} & \ms{0.597}{0.02} & \ms{0.516}{0.04} & \ms{0.629}{0.03} \\ \addlinespace[3pt]
    & RGF
    & \textbf{\ms{0.816}{0.01}} & \textbf{\ms{0.635}{0.03}} & \textbf{\ms{0.523}{0.02}} & \textbf{\ms{0.630}{0.02}}
    & \ms{0.800}{0.01} & \ms{0.615}{0.03} & \ms{0.509}{0.04} & \ms{0.616}{0.03}
    & \ms{0.790}{0.01} & \ms{0.615}{0.02} & \ms{0.510}{0.03} & \ms{0.618}{0.03} \\ \addlinespace[3pt]
    & TabNet
    & \textbf{\ms{0.777}{0.04}} & \textbf{\ms{0.533}{0.05}} & \textbf{\ms{0.471}{0.07}} & \textbf{\ms{0.597}{0.07}}
    & \ms{0.735}{0.02} & \ms{0.506}{0.05} & \ms{0.453}{0.07} & \ms{0.580}{0.07}
    & \ms{0.731}{0.02} & \ms{0.546}{0.01} & \ms{0.419}{0.08} & \ms{0.538}{0.08} \\
    \bottomrule
    \end{tabular}}%
    \end{table*}

    \subsection{Feature Interpretability}
    \label{sec:interpretability}

    To improve model transparency, we leverage the attention-based feature selection mechanism of the TabNet architecture~\cite{arik2021tabnet, rudin2019stop}, which provides a clinically interpretable summary of which variables contribute most strongly to recurrence-risk stratification. 
    Unlike tree-based ensembles that often require post-hoc analysis (e.g., SHAP), TabNet employs a sequential attention mechanism for automated feature interpretation~\cite{lundberg2017unified, arik2021tabnet}. Because many predictors are represented using categorical encodings, feature importance scores are aggregated across the corresponding one-hot encoded columns to recover importance at the level of the parent clinical attribute.

\begin{figure*}[htb]
        \centering
        \includegraphics[width=0.95\textwidth]{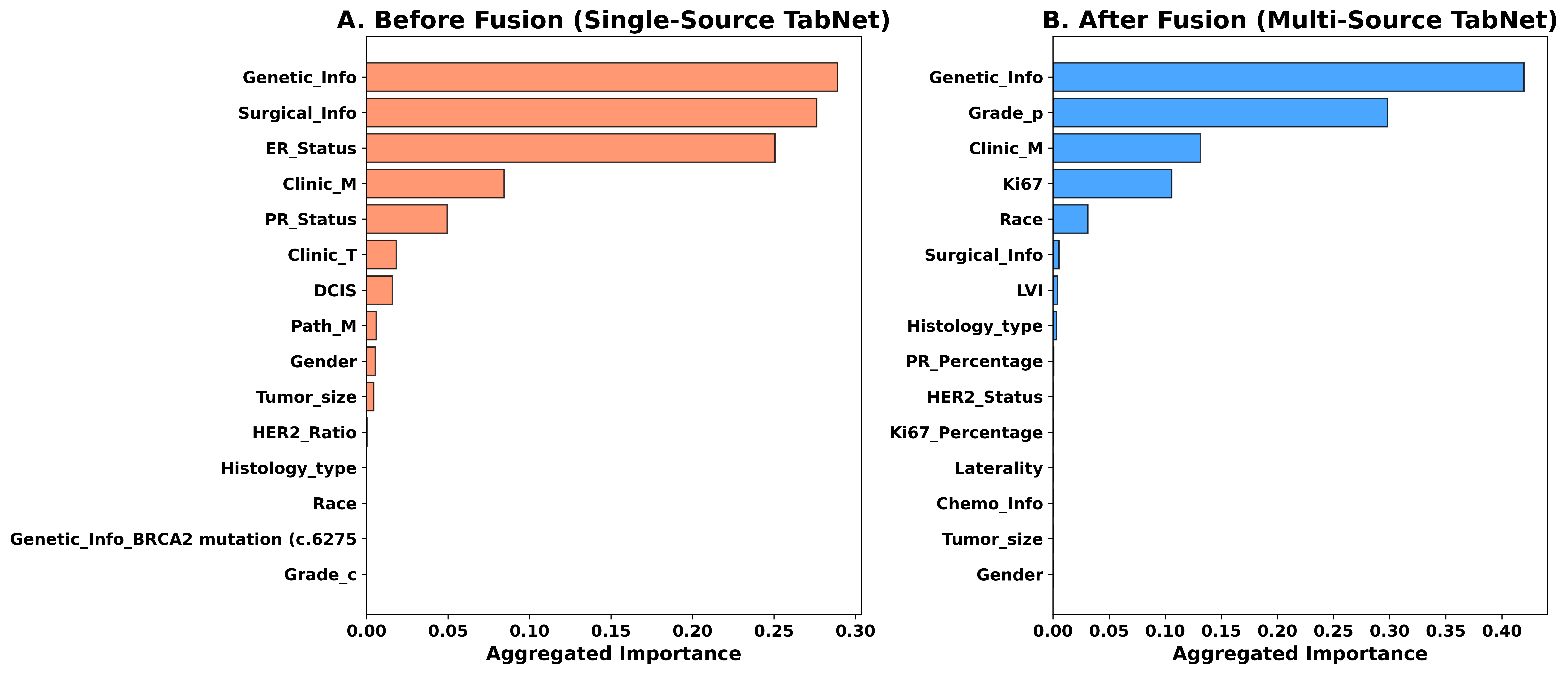} 
        \caption{Global feature importance (A) before and (B) after multi-source fusion using TabNet on the original imbalanced cohort. 
        }
        \label{fig:global_imp}
    \end{figure*}

    To evaluate whether the proposed reconciliation framework changes the learned evidence structure, we compare TabNet feature importance before and after multi-source fusion using the original imbalanced cohort. This comparison uses the same TabNet architecture in both settings, so the difference in feature importance reflects the effect of the input feature construction rather than a change in model class. As shown in Fig.~\ref{fig:global_imp}(A), before fusion, the single-source baseline primarily emphasizes variables already available in the original structured feature set, including genetic history, surgical information, ER and PR receptor status, and clinical staging variables. After multi-source fusion, Fig.~\ref{fig:global_imp}(B) shows that the feature-importance profile shifts toward harmonized and pathology-derived signals, with genetic history, pathologic grade, clinical metastasis status, Ki-67, surgical information, and LVI-related information appearing among the leading predictors. This shift indicates that multi-source harmonization changes the information used by the model: variables that were previously missing, fragmented, or inconsistently recorded become available in a more complete form and receive higher attention in the recurrence prediction task.
        
    We further assess the clinical plausibility of the model's data-driven logic by comparing the top-ranked features after data fusion against established clinical evidence. The high importance of genetic history is clinically plausible, as hereditary factors such as pathogenic variants can affect breast cancer risk and recurrence-related surveillance considerations~\cite{kuchenbaecker2017risks, malone2010population}. The emergence of pathologic grade among the leading after-fusion predictors is consistent with the established prognostic role of histologic grade in breast cancer outcomes~\cite{rakha2010breast, takahashi2020molecular}. The prominence of clinical metastasis status is also expected because metastatic staging is a core component of breast cancer prognostic staging and reflects disease extent~\cite{amin2017ajcc, giuliano2017breast}. Ki-67 is clinically meaningful because it reflects tumor proliferative activity and has been associated with relapse risk and survival outcomes in breast cancer~\cite{de2007ki, davey2021ki}. The prominence of LVI-related information is consistent with AJCC guidance that recognizes lymphovascular invasion as an adverse prognostic factor for metastasis~\cite{amin2017ajcc}. The appearance of surgical procedure among the important predictors is also clinically interpretable because local recurrence risk may differ by intervention type, margin status, and adjuvant therapy adherence~\cite{fisher2002twenty, agarwal2014effect}. Finally, to supplement the global importance analysis, we provide an instance-level attention-mask case study in Appendix~\ref{app:local_interpretability}. 
            
\section{Conclusions}
    This paper proposes a multi-modal harmonization framework for breast cancer recurrence prediction using heterogeneous clinical data. By leveraging regular expression extraction and a rigorous precedence-based fusion strategy, the proposed pipeline effectively addresses the challenges of data incompleteness and inconsistency inherent in single-source EHRs. Specifically, the integration of unstructured pathology narratives with structured registry and treatment records recovers definitive prognostic variables, such as TNM staging and biomarkers, thereby significantly reducing feature missingness without reliance on aggressive imputation. Experimental results demonstrate that this multi-modal approach enhances data quality and yields stronger discriminative performance than single-source baselines across multiple machine learning architectures, achieving higher AUROC and AUPRC scores in both controlled imbalance settings and the original imbalanced cohort.   
    
    We acknowledge that certain granular risk factors, specifically surgical margin status and details regarding the appropriateness of adjuvant therapy (e.g., radiation adherence following lumpectomy), were not explicitly harmonized in this iteration due to complexities in temporal logic and narrative variability. Future work will focus on developing specialized extraction logic for these variables to further refine the personalized risk profiles. Our framework provides clinicians with a scalable, reliable foundation for personalized risk stratification and survivorship care planning. Additionally, the proposed methodology for reconciling discordant clinical streams can be broadly applicable to other oncology domains where data fragmentation hinders effective modeling.

\appendix
\section{Supplementary Materials}
\label{app:supplementary}

\subsection{Variable Definitions}
\label{app:variables}
    Table \ref{tab:variables} provides an overview of the original span of the input variables for our dataset.

    \begin{table*}[htbp]
      \centering
      \caption{Variable definitions for the breast cancer dataset. Attributes marked with an asterisk (*) denote features included in the benchmark configuration based on prior literature \protect\cite{alzu2021predicting, gonzalez2023machine}.}
      \label{tab:variables}
    
      \begin{subtable}{\textwidth}
        \scriptsize
        \centering
        \caption{Patient demographics and staging variables.}
        \label{tab:vars_part1}
        \begin{tabularx}{\textwidth}{l>{\raggedright\arraybackslash}X>{\raggedright\arraybackslash}X}
          \toprule
          \textbf{Attribute} & \textbf{Description} & \textbf{Possible Values} \\
          \midrule
          Age*               & Patient age at diagnosis                                                        & Numeric values \\
          Race              & Self-reported race                                                              & White, Black, Asian, Native American, Pacific Islander, Hispanic/Latino \\
          Gender*            & Patient-reported gender                                                         & Female, Male, Not Stated \\
          Distance to care          & Distance from home to center                                                    & Numeric values \\
          Grade*             & Tumor grade (pathologic/clinical)                                                & Grade 1, Grade 2, Grade 3, GX \\
          Nodes Examined    & Total number of lymph nodes surgically removed and examined by pathology        & Numeric values \\
        Positive Nodes   & Number of surgically removed lymph nodes that tested positive for cancer cells  & Numeric values \\
          Tumor Size*        & Maximum tumor diameter                                                          & Numeric values (in mm) \\
          T*                 & Tumor stage (clinical/pathologic)                                             & TX, T0, Tis, T1, T2, T3, T4 \\
          N* (Pathologic)    & Pathologic nodal stage                                                          & pNX, pN0, pN1, pN2, pN3 \\
          N (Clinical)      & Clinical nodal stage                                                            & cNX*, cN0, cN1, cN2, cN3 \\
          M*                 & Metastasis status                                                               & MX, M0, cM1, pM1 \\
          DCIS*              & DCIS subtype                                                                    & Solid, Cribriform, Micropapillary, Comedo, Papillary, Apocrine, Mixed DCIS \\
          LVI*               & Lymphovascular invasion                                                         & Present, Not Present \\
          Histology Type    & Pathologic tumor classification                                                 & Invasive carcinomas, In situ carcinomas \\
          Surgical Procedure*     & Type of surgery performed                                                       & Mastectomy, Core Biopsy, Surgical Biopsy, Lumpectomy \\
          \bottomrule
        \end{tabularx}
      \end{subtable}
    
      \vspace{0.75em}
    
      \begin{subtable}{\textwidth}
        \scriptsize
        \centering
        \caption{Biomarkers, treatment, and genetic variables.}
        \label{tab:vars_part2}
        \begin{tabularx}{\textwidth}{l>{\raggedright\arraybackslash}X>{\raggedright\arraybackslash}X}
          \toprule
          \textbf{Attribute} & \textbf{Description} & \textbf{Possible Values} \\
          \midrule
          ER Status*         & Estrogen receptor status                                                        & Positive, Negative, Equivocal \\
          ER Percentage     & \% of ER-positive cells                                                         & Numeric values (\%) \\
          PR Status*             & Progesterone receptor status                                         & Positive, Negative, Equivocal \\
          PR Percentage         & \% of PR-positive cells                                              & Numeric values (\%) \\
          HER2 Status*           & HER2 receptor status                                                 & Positive, Negative, Equivocal \\
          HER2 Ratio (IHC/FISH) & HER2/CEP17 ratio by IHC or FISH                                     & Numeric values \\
          Ki-67 Status*          & Ki-67 proliferation index (categorical)                              & Low (\(<20\%\)), Intermediate (20--30\%), High (\(>30\%\)) \\
          Ki-67 Percentage      & Ki-67 proliferation index (\%)                                       & Numeric values (\%) \\
          Laterality*            & Tumor laterality                                                     & Unilateral, Bilateral, Multifocal, Multicentric \\
          Hormone Therapy*  & Hormone therapy status                                               & Administered, None, Refused, Contraindicated \\
          Immunotherapy    & Immunotherapy status                                                 & Administered, None, Refused, Contraindicated \\
          Chemotherapy           & Chemotherapy status                                                  & Administered, None, Refused, Contraindicated \\
          Genetic History          & Genetic mutation and family history                                  & Family history,  mutations in \textit{BRCA}/\textit{ATM}/\ldots \\
          Recurrence Outcome            & Recurrence outcome                                                   & Yes, No \\
          \bottomrule
        \end{tabularx}
      \end{subtable}
    \end{table*}

    \clearpage

\subsection{Sensitivity Analysis for Precedence-Based Reconciliation}
\label{app:precedence_sensitivity}
To evaluate whether the proposed hierarchy $\PATH \succ \ABS \succ \TS$ is a functional design choice rather than an arbitrary rule, we conduct a sensitivity analysis using an alternative fallback order, $\PATH \succ \TS \succ \ABS$. In both hierarchies, pathology reports remain the highest-priority diagnostic source. The alternative configuration reverses the secondary priority between $\ABS$ and $\TS$, prioritizing treatment-summary values before $\ABS$ values when pathology-derived values are unavailable. Because this analysis is conducted on the original imbalanced cohort, we interpret the results using threshold-independent metrics, i.e., AUROC and AUPRC.  Table~\ref{tab:precedence_sensitivity} shows that this alternative hierarchy generally reduces predictive performance relative to the proposed hierarchy. The decline is especially clear for TabNet, whose multi-source AUROC decreases to 0.779 under the alternative hierarchy, compared with 0.825 under the proposed hierarchy in the original imbalanced-cohort analysis. In addition, under the alternative hierarchy, the multi-source TabNet model does not outperform the single-source registry baseline in AUROC (0.779 vs. 0.808). These results support the use of ABS as the secondary source after PATH, because ABS provides more standardized and complete coded information than TS for many prognostic variables.

\begin{table*}[htb]
\centering
\scriptsize
\setlength{\tabcolsep}{4pt}
\renewcommand{\arraystretch}{0.95}
\caption{Sensitivity analysis using the alternative precedence hierarchy $\PATH \succ \TS \succ \ABS$ on the original imbalanced cohort. Single-source models use original features from the ABS modality. AUROC and AUPRC are reported as threshold-independent metrics. Best results for each algorithm across the three data sources are bolded.}
\label{tab:precedence_sensitivity}
\resizebox{0.85\textwidth}{!}{
\begin{tabular}{@{} l | cc | cc | cc @{}}
\toprule
\multirow{2}{*}{Model}
& \multicolumn{2}{c}{Multi-source}
& \multicolumn{2}{c}{Single-source}
& \multicolumn{2}{c}{Benchmark} \\
\cmidrule(lr){2-3} \cmidrule(lr){4-5} \cmidrule(lr){6-7}
& AUROC & AUPRC
& AUROC & AUPRC
& AUROC & AUPRC \\
\midrule
RF     & \textbf{0.818} & \textbf{0.258} & 0.806 & 0.249 & 0.778 & 0.185 \\
XGB    & \textbf{0.811} & \textbf{0.284} & 0.762 & 0.205 & 0.761 & 0.181 \\
RGF    & \textbf{0.831} & \textbf{0.284} & 0.817 & 0.245 & 0.816 & 0.274 \\
TabNet & 0.779 & 0.241 & \textbf{0.808} & \textbf{0.242} & 0.766 & 0.236 \\
\bottomrule
\end{tabular}
}
\end{table*}

\subsection{Instance-Level Interpretability}
\label{app:local_interpretability}

\begin{figure*}[htbp]
    \centering
    \includegraphics[width=1\textwidth]{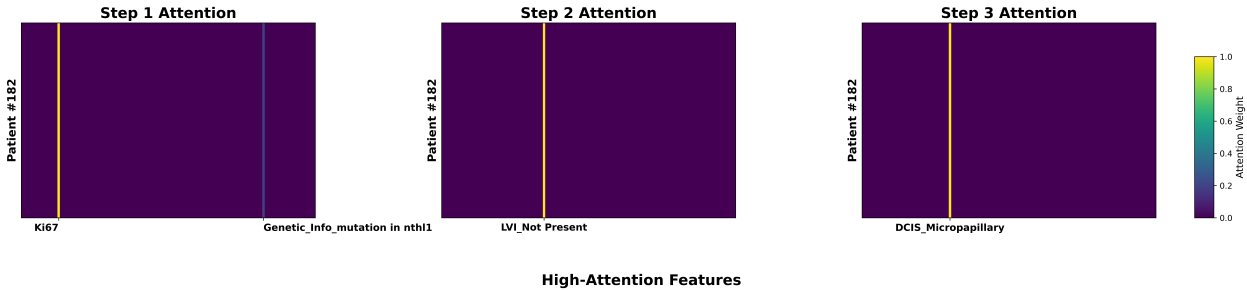}
    \caption{Instance-level attention pathways and feature prioritization for Patient 182. 
    }
    \label{fig:local_case}
\end{figure*}

To supplement the global feature-importance analysis in Section~\ref{sec:interpretability}, we provide an instance-level attention-mask visualization for a representative patient case. This analysis illustrates how TabNet sequentially selects and aggregates clinical variables when producing an individual recurrence-risk score. Fig.~\ref{fig:local_case} shows the attention pathways across decision steps for Patient 182. In the first decision step, the model assigns substantial attention to Ki-67-related information and genetic-variant information, indicating that proliferative activity and inherited-risk-related variables contribute to the initial risk assessment for this case. In the second decision step, the attention mask shifts toward LVI status, and in the third decision step, it shifts toward DCIS subtype. This sequential pattern illustrates how the model combines biomarker, genetic, and pathology-derived information when forming an individual prediction. 

\section*{Data Availability Statement}
This study was approved by the Institutional Review Board (IRB) at University of Tennessee Knoxville, approval ID \#5156. The study was conducted in accordance with relevant institutional guidelines and regulations. 
 The clinical data were sourced from the Research Enterprise Datawarehouse at the University of Tennessee Health Science Center. These data contain protected health information and are not publicly available due to institutional privacy policies and HIPAA regulations. De-identified data and the extraction logic may be made available to qualified researchers upon reasonable request to the corresponding author, subject to IRB approval and the execution of a Data Use Agreement.

\section*{Consent and approval}
This does not apply to this work as no human subjects were involved.

\section*{Disclosure statement}

No potential conflict of interest was reported by the authors.

\section*{Acknowledgment}
 This research has been supported by a seed grant from the Human Health and Wellness Gateway at the University of Tennessee, Knoxville.
 We acknowledge with deep gratitude Dr. John Bell, former Director of the Cancer Institute at the University of Tennessee Medical Center, whose mentorship, collaboration, and original initiation of this work were central to its development. His intellectual leadership and generosity continue to inspire us. This paper is dedicated to his memory.

\bibliographystyle{apacite}
\bibliography{ref}

\end{document}